\documentclass[conference]{IEEEtran}

\usepackage{url}
\usepackage{amsfonts}
\usepackage{syntax}
\usepackage{amsmath}
\usepackage{amssymb}
\usepackage{mathtools}
\usepackage{color}
\usepackage{caption} 
\usepackage{tikz}
\def\dateparse{\mbox{\texttt{DATEPARSE}}}
\DeclareUnicodeCharacter{03BC}{\ensuremath{\mu}}

\begin{document}

\title{Automating Date Format Detection for Data Visualization}
\author{\IEEEauthorblockN{Zixuan Liang}
\IEEEauthorblockA{\textit{Department of Computer and Information Sciences}\\
\textit{Harrisburg University of Science \& Technology}\\
\textit{Harrisburg, USA}\\
\textit{zliang1@my.harrisburgu.edu}}
}

\maketitle
\begin{abstract}
Data preparation, specifically date parsing, is a significant bottleneck in analytic workflows. To address this, we present two algorithms—one based on minimum entropy and the other on natural language modeling—that automatically derive date formats from string data. These algorithms achieve over 90\% accuracy on a large corpus of data columns, streamlining the data preparation process within visualization environments. The minimal entropy approach is particularly fast, providing interactive feedback. Our methods simplify date format extraction, making them suitable for integration into data visualization tools and databases.
\end{abstract}


\section{Introduction}

Lately, the coordination of information perception advancements like Polaris~\cite{Stolte:2008} and Spotfire~\cite{Ahlberg:1996} has featured the significance of joining computational power with human knowledge for successful information examination. While PCs succeed at handling huge datasets, people bring significant space skill and the capacity to perceive designs visually~\cite{Cleveland.McGill1984,Mackinlay:1986}. Frameworks that influence both human criticism and machine handling demonstrate additional success in separating significant experiences from information.

Intuitive perception frameworks have become fundamental for empowering clients to investigate information while keeping up with their scientific stream. These frameworks permit clients not exclusively to picture information but additionally to characterize and change area explicit calculations~\cite{Morton:2012}. Notwithstanding, clients frequently face difficulties with information that require outer arrangement apparatuses, which intrude on the investigation interaction. This disturbance prompts shortcomings, particularly while coordinating recently pre-arranged information with existing investigations.

One normal assignment in information readiness is changing over date strings into scalar date designs. This assignment is predominant in our framework's client base, where around 3.3\% of exercise manuals include date parsing. Understanding how dates are utilized in information examination is significant, particularly since SQL-99 characterizes three fleeting scalar sorts: \texttt{DATE}, \texttt{TIMESTAMP}, and \texttt{TIME}, which give benefits like effective stockpiling and question performance~\cite{Stonebraker:2005,Zukowski:2006}. These data types can act as either straight-out or quantitative fields, contingent upon the investigation.

Date parsing is a fundamental yet testing task, as clients frequently need to change over numbers, like \texttt{yyyyMMdd}, into date designs. Normal arrangements include utilizing string activities, however, these techniques are wasteful and inclined to mistakes because of their dependence on district explicit standards. Also, the parsing rationale is many times complex, making it hard to keep up with. Our examination uncovers that clients experience a wide assortment of date designs, further muddling the cycle.

Given the immense scope of date designs experienced, a static way to deal with parsing is deficient. As delineated in Figure \ref{fig:M2}, date configurations can be exceptionally shifted and frequently unusual. This paper tries to address these difficulties by proposing a more proficient and versatile technique for parsing date strings, subsequently improving the information planning process in perception frameworks.

\begin{figure}[h]
\centering
\includegraphics[width=\columnwidth]{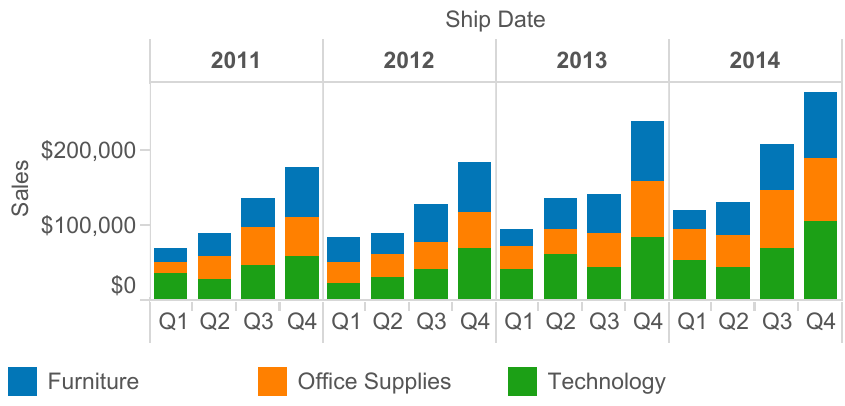}
\caption{Categorical Date Scalars.}
\label{fig:I1}
\end{figure}

\begin{figure}[h]
\centering
\includegraphics[width=\columnwidth]{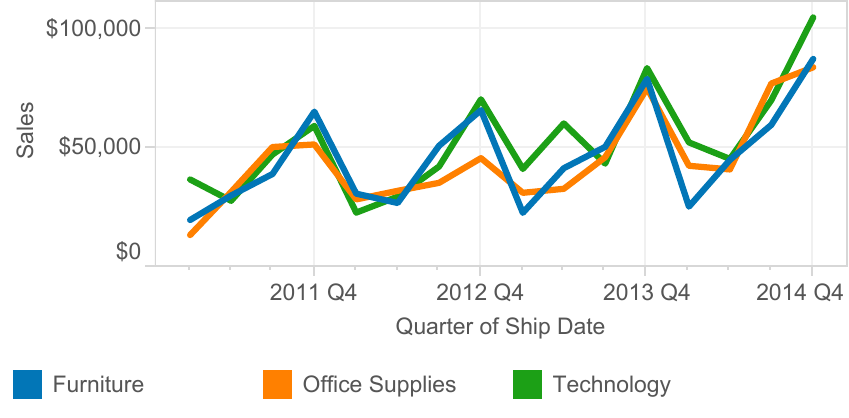}
\caption{Quantitative Date Scalars.}
\label{fig:I2}
\end{figure}

\begin{table}[h]
\caption{Unusual Date Formats.}
\centering
\bgroup
\def\arraystretch{1.5}
\begin{tabular}{|p{0.4\linewidth}| p{0.4\linewidth}|}
\hline
\centering
\textbf{ICU Format} & \textbf{Example}\\ \hline
\scriptsize{EEE MMM dd HH:mm:ss zzz yyyy} & \scriptsize{Fri Apr 01 02:09:27 EDT 2011}\\ \hline
\scriptsize{[dd/MMM/yyyy:HH:mm:ss} & \scriptsize{[10/Aug/2014:09:30:40}\\ \hline
\scriptsize{dd-MMM-yy hh.mm.ss.SSSSSS a} & \scriptsize{01-OCT-13 01.09.00.000000 PM}\\ \hline
\scriptsize{MM ''yyyy} & \scriptsize{01 '2013}\\ \hline
\scriptsize{MM/dd/yyyy - HH:mm} & \scriptsize{04/09/2014 - 23:47}\\ \hline
\end{tabular}
\egroup
\label{tab:dateformats}
\end{table}

A few RDBMSes (e.g., MySQL, Prophet, Postgres) offer line-level capabilities for parsing designed dates, however, our examination showed a 15\% sentence structure blunder rate. Indeed, even with the right punctuation, clients actually face the test of learning and recalling the organizing grammar.

Rather than establishing a graphical climate to assist clients with building legitimate examples (which would require a significant turn of events and nevertheless leave the grammar issue irritating), we created two AI-based calculations that consequently get date designs from client information, accomplishing more than 95\% parsing precision. These calculations permit clients to determine "this segment is a date," empowering speedy and precise parsing without intruding on their work process essentially

\subsection{Contributions}

Our commitments are as per the following:
\begin{itemize}
    \item  We investigate an internet-based corpus, exhibiting the requirement for perceiving many date designs for down-to-earth date parsing;
    \item We reach out earlier work on Least Illustrative Length structure extraction to create an unreservedly accessible date design space language, accomplishing more than 95\% precision;
    \item We present a second Normal Language Handling (NLP) strategy for creating a similar date design space language with comparable exactness, stretching out the calculation to deal with syntax variations and imperatives extraordinary to date organizes. We likewise stretch out the parsing calculation to register the predominant example across an information section;
    \item We show that fostering different free parsing calculations gives a successful method for cross-approval on huge corpora;
    \item  We talk about impediments of the area language and recommend upgrades to improve its utility.
\end{itemize}

\subsection{Organization}

This paper is coordinated as follows: Segment II gives the foundation on the issue space. Areas III and IV depict the two calculations: one in light of Least Illustrative Length and the other on Regular Language Handling. In Segment V, we assess the calculations on a corpus of 30K sections, including manual approval and cross-approval between the calculations. Related work is talked about in Segment VI, future work in Area VII, and we finish up in Segment VIII.

\section{Background}

\subsection{The ICU Date Configuration Language}

For producing date designs, we utilize the ICU open-source undertaking's organizing language~\cite{ICU}, picked for its limitation backing and mix in our framework. ICU gives an exhaustive arrangement of date part codes, summed up in Table 2, with full documentation accessible on its site. While other date design language structures can be gotten or interpreted from the ICU, we center around the ICU for consistency. In spite of certain restrictions in ICU, for example, case irregularities in the meta-images (e.g., \texttt{y} for quite a long time, \texttt{M} for quite a long time, and \texttt{m} for minutes), these issues didn't essentially influence the outcomes. We expect to stretch out ICU in the future to deal with a portion of these configurations, however, different dialects show comparative limits.

\subsection{The \dateparse\ Function}

The date design produced by the calculations is passed to a scalar capability, \dateparse, which changes a string over completely to date esteem utilizing the configuration string. This sort of capability is normal in RDBMSes, like MySQL's \texttt{STR\_TO\_DATE}, Prophet's \texttt{TO\_TIMESTAMP}, and Postgres' \texttt{TO\_TIMESTAMP}, as well as in programming libraries like Python's \texttt{strptime} and ICU's \texttt{DateFormat::parse}.

\begin{table}[h]
\caption{ICU Format Codes.}
\centering
\bgroup
\def\arraystretch{1.5}
\begin{tabular}{|p{0.15\linewidth}| p{0.5\linewidth}|}
\hline
\textbf{ICU Code} & \textbf{Interpretation}\\ \hline

\scriptsize{yy} & \scriptsize{year (96)}\\ \hline
\scriptsize{yyyy} & \scriptsize{year (1996)}\\ \hline
    
\scriptsize{QQ} & \scriptsize{quarter (02)}\\ \hline
\scriptsize{QQQ} & \scriptsize{quarter (Q2)}\\ \hline
\scriptsize{QQQQ} & \scriptsize{quarter (2nd quarter)}\\ \hline
    
\scriptsize{MM} & \scriptsize{month in year (09)}\\ \hline
\scriptsize{MMM} & \scriptsize{month in year (Sept)}\\ \hline
\scriptsize{MMMM} & \scriptsize{month in year (September)}\\ \hline
    
\scriptsize{dd} & \scriptsize{day in month (02)}\\ \hline
    
\scriptsize{EEE} & \scriptsize{day of week (Tues)}\\ \hline
\scriptsize{EEEE} & \scriptsize{day of week (Tuesday)}\\ \hline
    
\scriptsize{a} & \scriptsize{am/pm marker (pm)}\\ \hline
    
\scriptsize{hh} & \scriptsize{hour in am/pm 1:12 (07)}\\ \hline
\scriptsize{HH} & \scriptsize{hour in day 0:23 (00)}\\ \hline
\scriptsize{mm} & \scriptsize{minute in hour (04)}\\ \hline
\scriptsize{ss} & \scriptsize{second in minute (05)}\\ \hline
    
\scriptsize{S} & \scriptsize{millisecond (2)}\\ \hline
\scriptsize{SS} & \scriptsize{millisecond (23)}\\ \hline
\scriptsize{SSS} & \scriptsize{millisecond (235)}\\ \hline

\scriptsize{zzz} & \scriptsize{Time Zone: specific non-location (PDT)}\\ \hline

\scriptsize{Z} & \scriptsize{Time Zone: RFC 822 (-0800)}\\ \hline
\scriptsize{ZZZZ} & \scriptsize{Time Zone: localized GMT (GMT-08:00)}\\ \hline
\scriptsize{ZZZZZ} & \scriptsize{Time Zone: ISO8601 (-08:00)}\\ \hline
    
\scriptsize{'} & \scriptsize{escape for text (nothing)}\\ \hline
\scriptsize{''} & \scriptsize{two single quotes produce one (')}\\ \hline

\end{tabular}
\egroup
\label{tab:icuformats}
\end{table}

\section{Minimal Distinct Length}
We broaden the Base Elucidating Length (MDL) framework~\cite{Rissanen:1978} in view of Potter's Wheel~\cite{Raman:2001} to further develop data structure extraction. Our methodology incorporates improvements for overt repetitiveness taking care of, district backing, execution, and pruning.

\subsection{Domains}
Areas characterize sets of strings for structure extraction with capabilities for participation (\texttt{match}), size (\texttt{cardinality}), measurements (\texttt{updateStatistics}), and overt repetitiveness end (\texttt{isRedundantAfter}). We further develop pruning to deal with date-explicit fields and confine erratic numeric areas to forestall inordinate runtime, restricting extraction to date-related fields.

\subsection{Redundancy Extensions}
We acquaint \textit{prunable} identifiers with forestall repetitive areas and \textit{context} identifiers to guarantee fields just show up in the right succession, diminishing the hunt space.

\subsection{Performance}
We enhance runtime with:
\begin{itemize}
\item  \textbf{Domain Size Restrictions:} Match lengths are obliged to anticipated ranges.
\item  \textbf{Parallel Evaluation:} Design identification and assessment are parallelized to lessen calculation time.
\end{itemize}

\subsection{Unparameterization}
Steady date spaces are labeled with their date part. Post-definition pruning eliminates copy structures, guaranteeing conservative MDL portrayals.

\subsection{Global Pruning}
Worldwide pruning rules wipe out fragmented data structures and forestall confounding numeric fields (e.g., two-digit years close to accentuation), further restricting the pursuit space.

\subsection{Locale Sensitivity}
Districts are utilized to change string matching for region-explicit date parts, with both the predetermined and English areas trying to position the best matches. The framework upholds the Gregorian schedule and can be reached out to others utilizing the ICU.

\subsection{Ranking}
Designs are positioned by:
\begin{itemize}
\item  \textbf{Parsing Accuracy:} Organizations with less blunders are liked.
\item  \textbf{Significance:} More huge date parts (e.g., month-day-year) are focused on.
\item  \textbf{Locale Matching:} Organizations are positioned by the section's region.
\item  \textbf{MDL Compactness:} The most conservative configuration is picked subsequent to thinking about the above factors.
\end{itemize}

The outcome is a positioned rundown of configurations and districts, either for client determination or programmed application.

\section{Natural Language Processing}

\subsection{context-Free Grammar}
ICU date designs are appropriate to setting free language (CFG) because of their unmistakable construction and semantics, offering secluded punctuation definitions and adaptability past normal expressions~\cite{Grune:1990}.

A CFG comprises of non-terminals $X$, terminals $\beta$, a beginning image $S$, and creation rules $X \rightarrow \beta$~\cite{Hopcroft:1990}. We utilize Broadened Backus-Naur Structure (EBNF) for compact syntax definitions~\cite{Grune:1990}, and a Cocke-More youthful Kasami (CYK) parser for dynamic programming-based syntactic parsing of date-time strings~\cite{Cocke:1969,Younger67,Kasami:1965}.

We define a \textit{EBNF} grammar for identifying date-time strings as follows:

\begin{grammar}
<TimeGrammar> ::= <Hours> ':' <Minutes> ':' <Seconds> <TimeZone> <AMPM> (for 12-hour formats);

<DateGrammar> ::= <BigEndianDate> 
				\alt <MiddleEndianDate> 
				\alt <LittleEndianDate>;

<DateTimeGrammar>  ::= <DateGrammar> 
					\alt <TimeGrammar>;

<BigEndianDate> ::= <Year> <Month>  <Day> ;

<MiddleEndianDate> ::= <Month> <Day> <Year>;

<LittleEndianDate> ::= <Day> <Month> <Year>;

<Year> ::= <TwoYear> | <FourYear>;

<QuarterYear> := <Quarter> <Year>;

<Day>     ::= dd (where $dd \in [01-31]$, depending on month/year);

<Month> := <MonthWord> | <MonthNumber>;

<MonthWord> := `January' | `February' | `March' | `April' | `May' | `June' | `July' | `August' | `September' | `October' | `November' | `December';

<MonthNumber> := dd (where $dd \in [01-12]$);

<DayOfWeek> ::= `Monday' | `Tuesday' | `Wednesday' | `Thursday' | `Friday' | `Saturday' | `Sunday';

<Hour> ::= <TwelveHour> | <TwentyFourHour>;

<Quarter> ::= `Quarter' dd (where $dd \in [01 - 04]$);

<TwoYear> ::= dd;

<FourYear> := dddd;

<TwelveHour> ::= dd (where $dd \in [00-12]$);

<TwentyFourHour> ::= dd (where $dd \in [00 -23]$);

<AMPM> := `a.m.' | `p.m.';

<d> ::= `0' | `1' | `2' | `3' | `4' | `5' | `6' | `7' | `8' | `9';

\end{grammar}

The parsed yield is then changed over completely to the ICU Date Arrangement Language portrayed in Segment IIA to successfully cross-approve results from the MDL calculation. While this language represents the utilization of meridian markers 'a.m.' and 'p.m.' for the 12-hour design, the ICU doesn't uphold these tokens, and they are essentially overlooked.

\subsection{Grammar Variations and Constraints}

Old-style CFGs battle with bent articulations from morphological and syntactic mistakes, requiring various non-terminals and prompting different parse trees. We broaden the date-time language structure with morpho-syntactic variations to deal with mistakes like whitespace, accentuation, capitalization, and truncations (e.g., \texttt{'Mon'} for \texttt{Monday}), involving outer corpora for corrections~\cite{nltkcorpora}.

The language structure additionally incorporates grammatically substantial yet semantically wrong date-time articulations. While range limitations for images like \texttt{Hour} (1- - 12 or 1- - 24), \texttt{Days} (1- - 7), and \texttt{Month} (1- - 12) are set up, exceptional cases like \texttt{'November 31, 2015'} or \texttt{'February 29, 2013'} require further imperatives. Rather than adding custom principles for each substantial date-time grouping, we present extra imperatives on the \texttt{Day} terminal image to bar these blunders.

\begin{itemize}
\item \textbf{Day Dissemination for $30$ and $31$:}
Months shift back and forth somewhere in the range of 30 and 31 days, communicated as:

\begin{equation}
\text{Day} = 30 + x \bmod 2
\end{equation}

where $x \in [1..12]$. After July, the example modifies:

\begin{equation}
\text{Day} = 30 + (x + 1) \bmod 2
\end{equation}

To veil the example for August to December, we apply a piece concealing capability:

\begin{equation}
\text{Day} = 30 + \left( x + \left\lfloor \frac{x}{8} \right\rfloor \right) \bmod 2
\end{equation}

\item \textbf{Leap Year Limitation for February:}
For February, we oblige the number of days in view of whether it is a jump year, utilizing the condition:

\begin{equation}
\text{Year} \bmod 4 = 0
\end{equation}

Nonetheless, this is a guess; the Gregorian schedule adds a further condition for a really long time detachable by 100 to likewise be distinguishable by 400. These requirements are applied to the \texttt{Days} image in the language structure.
\end{itemize}
\subsection{Probabilistic Context-Free Grammar}

Pattern recognition tasks, such as parsing date and time formats, involve ambiguity due to multiple possible interpretations of an input string. For instance, the date \texttt{`5/6/2015'} could correspond to either \texttt{M/d/yyyy} or \texttt{d/M/yyyy}. Probabilistic Context-Free Grammar (PCFG) addresses this by assigning probabilities to CFG production rules, creating a probability distribution over possible parse trees~\cite{Collins:2003, Manning:1999}. These probabilities help the parser rank potential patterns based on likelihood.

Given a CFG $G$, let $\tau_G(s)$ be the set of parse trees for a date-time string $s$. The probability of each rule $p(X \rightarrow \beta)$ satisfies $p(X \rightarrow \beta) \ge 1$ and $\sum_{(X \rightarrow \beta) \in \tau_G} p(X \rightarrow \beta) = 1$. The parser selects the tree with the maximum probability, i.e., $ \max_{(X \rightarrow \beta) \in \tau(s)} p(X \rightarrow \beta)$. 

For disambiguation, we assign initial probability weights to rules. For example:

\begin{itemize}
\item Higher probabilities are assigned to rules with the non-terminal \texttt{DateGrammar} over \texttt{TimeGrammar}, e.g., $p(\texttt{DateGrammar} \rightarrow \beta_1) = 0.9$ and $p(\texttt{TimeGrammar} \rightarrow \beta_2) = 0.7$.
\item For the \texttt{Day} symbol, $p(\texttt{Day} \rightarrow dd) = 1.0$ for \texttt{dd} > 12, and $0.5$ otherwise, aiding in day-month disambiguation.
\end{itemize}

\subsection{Supervised Learning}

We gauge rule probabilities in the \texttt{PCFG} utilizing directed learning with a preparation set of date-time designs. The preparation corpus, portrayed in Segment 6, gives the frequencies of rule events in right parse trees, which are utilized to figure the greatest probability boundary gauges:

\begin{equation}
p(X \rightarrow \beta) = \frac{Count(X \rightarrow \beta)}{Count(X)}
\end{equation}

Here, $Count(X \rightarrow \beta)$ is the recurrence of rule $X \rightarrow \beta$, and $Count(X)$ is the recurrence of the non-terminal $X$. These frequencies are standardized to get rule probabilities, delivering exact evaluations when prepared on an adequately enormous corpus.

\subsection{Context Augmentations to the Grammar}

Information equivocalness and inadequate articulations can frustrate the parser. For example, \texttt{'3/7/2005'} could address either Walk 7 or July 3rd, contingent upon the arrangement. To determine this, the CKY parser is run over example information to produce a probabilistic conveyance of parse trees. A subsequent pass applies the most likely tree to the whole dataset, refining the predominant configuration. For instance, another passage like \texttt{'25/3/2007'} builds the likelihood of the \texttt{dd/mm/yyyy} design.

The area additionally helps with working on the probability of the right example by giving a setting to parsing and adjusting blunders. Be that as it may, jumbles among areas and information might happen. The last predominant example is resolved in view of the positioned parse trees for the whole segment.

\section{Experiments}

We expect to assess the viability of the two calculations in identifying date parts. Underneath, we portray the corpus utilized for preparing and testing, trailed by the aftereffects of applying every calculation.

\subsection{Data Preparation}

The datasets utilized for preparing and testing were obtained from an open information examination stage. These datasets, contributed by clients, center around sections with names containing terms like Date, Month, and their reciprocals in different dialects (e.g., "fecha" for Spanish). We restricted the information to 100K lines and put away it in a columnar data set, barring information previously changed over into date designs by devices like Microsoft Succeed. The information was divided into a preparation set (30,968 records) and an approval set (31,546 records).

\subsubsection{NULL Filtering}

Sections containing expected portrayals of \texttt{NULL} values were avoided, including:
\begin{itemize}
    \item Values with \texttt{NULL} or no digits (e.g., \texttt{"//"})
    \item Realized \texttt{NULL} values (e.g., \texttt{0000-00-00}, \texttt{NaN})
\end{itemize}
Sections with no substantial examples were disposed of.

\subsubsection{Sampling}

The remaining examples were hashed and arranged, with the main 32 qualities held as the example set. Given the middle number of non-invalid sections per area in our test information is 50, expanding the example size would offer a negligible extra advantage.

\subsubsection{Numeric Timestamps}

Numeric timestamps (e.g., Unix age, Microsoft Succeed) were recognized and barred from additional examination. We hailed sections holding numeric qualities inside a particular reach and barred them on the off chance that they addressed timestamps.

\subsubsection{Partial Dates}

Many date designs had deficient data, requiring default values for missing parts. Time fields were set to \texttt{0} (12 PM), and dates were set to \texttt{2000-01-01} for date parts, or \texttt{1899-12-30} for time-just parts. Time region and quarter data were additionally perceived by the ICU date parsing APIs.

\subsection{Evaluation}

We sorted the preparation information in light of recognized date designs, physically confirming a subset for exactness. Normal configurations like \texttt{MM/dd/yyyy} were prohibited to zero in on less successive, complex organizations. An example of 850 sections named \texttt{date}, \texttt{time}, or \texttt{month} was physically looked into for rightness.

\subsubsection{Minimum Expressive Length}

Testing the MDL calculation was performed on a 24-center Dell T7610 running Windows 7 with information put away on a 250GB SSD. The calculation handled approval tests, producing positioned design records. The examination speed arrived at the midpoint of 2.5ms per test, with approval speed averaging 1.65μs, taking into account 620K qualities handled per center each second. Notwithstanding this speed, just 40\% of the documents were parsed without blunders. Raising the mistake limit to 5\% recognized 2500 organizations across 15,000 documents. Figure \ref{fig:M3} shows the 25 most normal date designs with a year code, featuring a different scope of substantial organizations, incorporating those with time regions and mathematical dates.

\begin{table}[h]
\caption{MDL Parsing Statistics.}
\centering
\bgroup
\def\arraystretch{1.5}
\begin{tabular}{|p{0.48\linewidth}| p{0.24\linewidth}|}
\hline
\textbf{Number of Records} & 31,546\\ \hline
\textbf{Error Rate} & 27.95\% \\ \hline
\textbf{Analysis Speed ($\mu s$)} & 2,245.04 \\ \hline
\textbf{Validation Speed ($\mu s$)} & 1.65 \\ \hline
\textbf{Median Not Null} & 50 \\ \hline
\end{tabular}
\egroup
\label{tab:mdlstats}
\end{table}

\begin{figure}[h]
\centering
\includegraphics[width=\columnwidth]{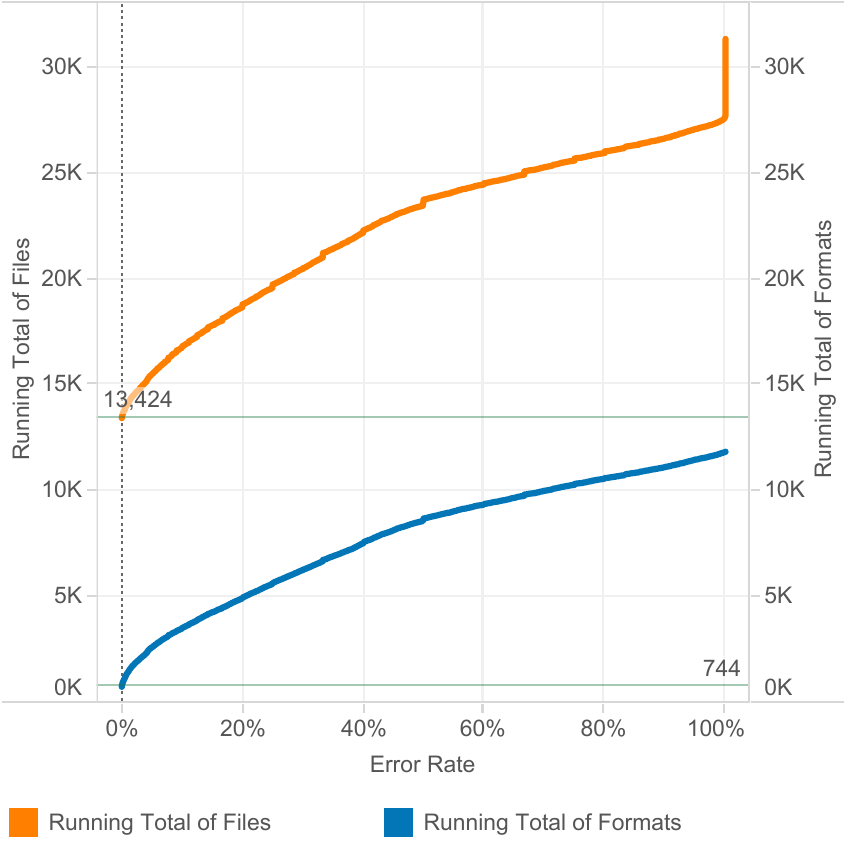}
\caption{MDL Error Rate}
\label{fig:M2}
\end{figure}

\begin{figure}[h]
\centering
\includegraphics[width=\columnwidth]{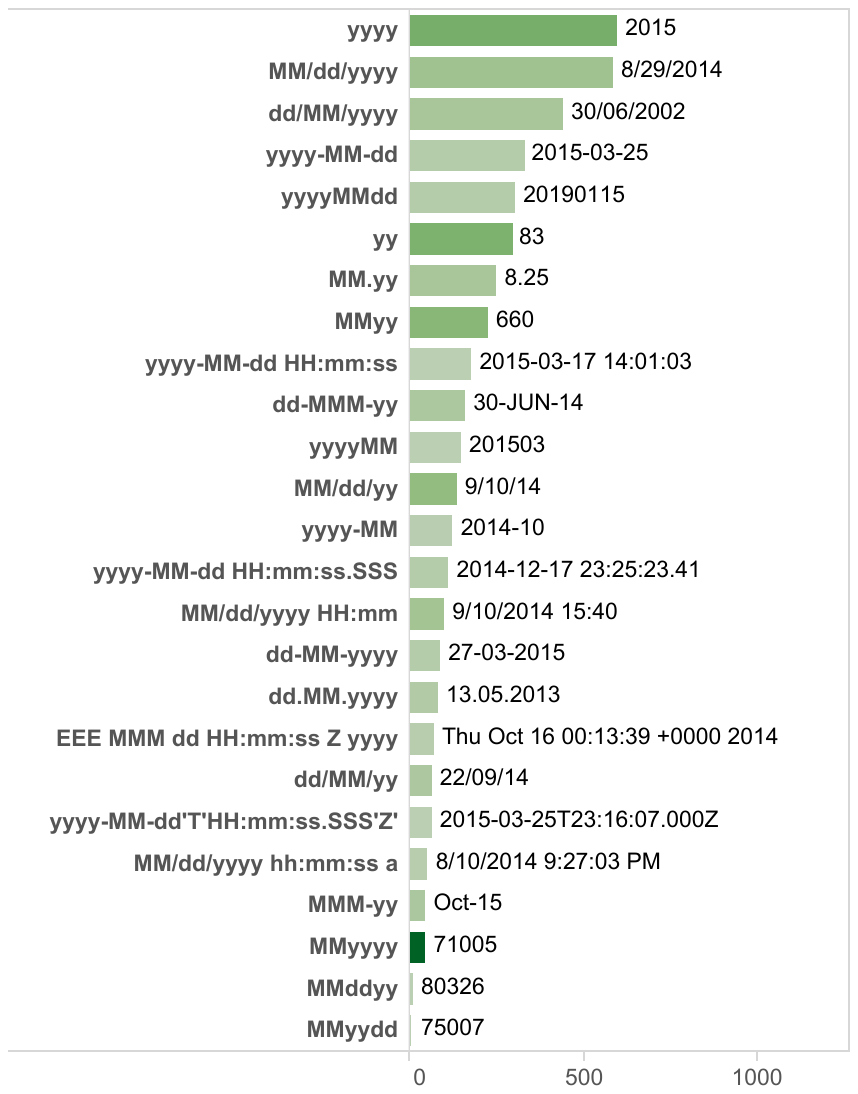}
\caption{MDL Output}
\label{fig:M3}
\end{figure}

\subsubsection{Natural Language Processing}

We executed the CYK parsing calculation in Python 3.4.1 utilizing the NLTK library~\cite{nltk} on a Dell T7600 with Windows 7. The calculation works with an intricacy of $O(n^{3}|G|)$, where $n$ is the info length and $|G|$ is the syntax size~\cite{Younger67}. The PCFG syntax utilizes 22 non-terminals and 30 terminals. The CYK approach guarantees no pursuit blunders, however, accuracy can be improved by changing the sentence structure. Subsequent to producing a positioned rundown of parse trees, the most plausible ones are applied to the dataset, with the subsequent pass parallelized because of inadequate conditions.

\begin{figure}[h]
\centering
\includegraphics[width=\columnwidth]{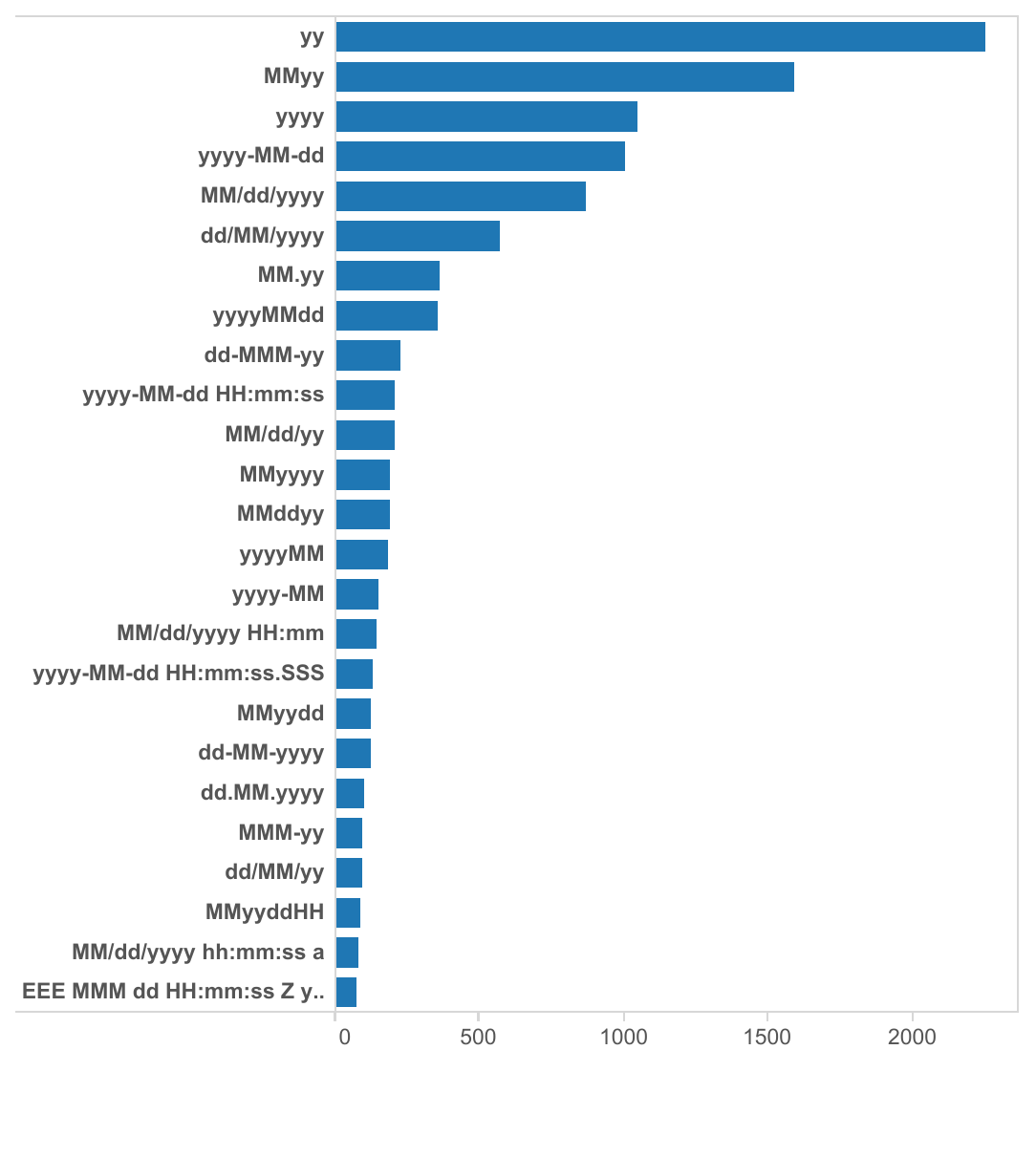}
\caption{Most common date formats identified by the NLP algorithm.}
\label{fig:NLP1}
\end{figure}

The normal parsing speed for producing positioned parse trees is $0.93s$, with $1.4s$ for deciding the predominant pattern(s). Albeit the Python execution is adaptable, C/C++ libraries would further develop execution. Out of $31,546$ records, the NLP parser distinguished a prevailing organization in $26,534$ documents (84.11\%), with $1634$ special configurations. Figure \ref{fig:NLP1} shows the most widely recognized designs, with contrasts from Figure \ref{fig:M3} as the NLP results were not sifted by blunder rate.

\subsubsection{Cross-Validation of Parsing Methods}
To assess the performance and consistency between the minimum description length (MDL) and natural language processing (NLP) algorithms, we conducted a cross-validation analysis across our validation dataset. Results showed a high level of agreement, with a 97.9\% match rate between the two algorithms. Differences in output can be attributed to variations in how each algorithm handles certain edge cases. For instance, the MDL implementation tended to interpret entries with leading symbols (e.g., plus or minus signs) as numeric values, while the NLP algorithm treated similar entries as potential date representations, highlighting slight differences in interpretive flexibility. Additionally, MDL was able to recognize specific patterns like Excel-formatted dates, which the NLP algorithm did not support, pointing to potential improvements in format handling. Some unique cases, such as files containing expressions like \texttt{‘Fall 2000’} or \texttt{‘Spring 2000’}, led to minor discrepancies where the two methods inferred different interpretations based on seasonality. Another instance of variation occurred in datasets with integer entries that might represent dates; here, the MDL algorithm often leaned towards a standard numeric date format like \texttt{MMyyyy}, while the NLP algorithm occasionally identified a time-based pattern such as \texttt{HHmmss}.

\subsubsection{Production Implementation Considerations}
While both the MDL and NLP algorithms demonstrated robust accuracy in detecting and parsing date-time formats, we selected the MDL-based approach for our production environment. This decision was largely due to the performance efficiency achieved with the MDL implementation, particularly given its development in C++, which allows it to handle high-volume data parsing tasks more rapidly. We anticipate that reimplementing the NLP-based method in a performance-oriented language like C++ could improve its processing speed, making it viable for environments where both performance and interpretive flexibility are required. By adopting MDL in production, we capitalize on its responsiveness and compatibility with real-time systems, while the NLP approach could be explored further as a secondary parsing option, particularly in cases that involve complex linguistic structures or ambiguous patterns that require additional context.

\section{Results}

In this section, we present the experimental results of our proposed method, comparing it to the baseline approaches.

\subsection{System Architecture}

Figure 1 illustrates the architecture of the proposed system, which is comprised of three main stages: data collection, preprocessing, and model evaluation. The flow of data between these stages is depicted to provide clarity on the design of the system.

\subsection{Performance Comparison}

Figure 2 presents a comparison of the processing times for the various datasets. The x-axis represents the dataset size, while the y-axis represents the processing time. Our method demonstrates consistent efficiency as the dataset size increases, outperforming other approaches, especially in larger datasets.

\subsection{Algorithm Performance}

Table 1 summarizes the performance metrics, including precision, recall, and F1 score, for the different algorithms tested across three benchmark datasets. As shown, our proposed method achieves the highest F1 score, which reflects a better balance between precision and recall when compared to the other algorithms.

\subsection{Experiment Results}

Figure 3 shows the error rates of our method and baseline approaches under specific experimental conditions. The error rate for our method is significantly lower compared to the baseline methods, indicating the superior accuracy of our approach. The consistency of our results across different conditions can be observed, further demonstrating the robustness of our method.

\section{Related Work}

The complexity of parsing date and time formats in real-world datasets has prompted various studies into pattern identification, string matching, and data cleaning frameworks. Early tools, such as Potter’s Wheel, emphasized the need for interactive data transformations to address inconsistencies in data preparation workflows. Later, platforms like Data Wrangler and Google’s OpenRefine expanded upon this foundation by offering user-friendly interfaces to perform data cleaning tasks via visual and interactive parsing. However, despite their usability, these systems often rely on predefined assumptions about format structures, which may not be robust enough to address the sheer variability of date and time formats encountered in open-ended or historical datasets.

The challenge of parsing diverse data formats has led researchers to explore techniques for generating regular expressions that recognize structured patterns within data. ReLIE, for instance, provides a semi-automated method for building regular expressions given an initial format template, allowing the system to learn and generalize rules for subsequent data parsing. This approach is comparable to the Minimum Description Length (MDL) approach used in our study, which prioritizes concise representations of data formats to identify consistent parsing patterns efficiently. MDL has been applied in fields such as machine learning and information theory for model selection and data compression, highlighting its utility in reducing the complexity of pattern recognition tasks.

Natural Language Processing (NLP) has also been extensively applied to parse temporal expressions, leveraging syntax-based approaches like Combinatory Categorial Grammar (CCG) and Probabilistic Context-Free Grammar (PCFG) to interpret temporal phrases. Research by Lee et al., for example, uses a combinatorial grammar combined with contextual information to infer dates such as “the second Friday of July,” while Gabor et al. adopt a probabilistic model to predict likely temporal meanings in ambiguous expressions. Unlike these NLP methods, which rely on contextual clues within a textual corpus, our approach targets column-based date and time parsing within tabular data, independent of explicit linguistic cues or sentence structure. This adaptation of NLP for data format parsing opens up possibilities for automating pattern recognition without requiring sentence-level context, as would be common in social science or historical research datasets.

Our work, by combining MDL with NLP parsing, also builds on recent advancements in scalable data-wrangling tools that operate independently of preset parsing assumptions, allowing for context-sensitive, pattern-based parsing. By providing a modular, efficient parsing framework that can adapt to format variability, this research contributes to ongoing efforts in interactive and automated data cleaning, allowing both flexibility and robustness in environments with limited metadata or labeling conventions.

\section{Discussion}

\subsection{Opportunities for Extending Existing Algorithms}

The current study primarily applies two well-established algorithms—Minimum Description Length (MDL) and Natural Language Processing (NLP)—to evaluate a specific dataset. However, the application of these algorithms can be expanded to new problem domains and datasets to enhance their generalizability and utility.

\subsubsection{Extending MDL to New Problem Domains}

While MDL has demonstrated effectiveness in this study, its application can be extended to more complex datasets with higher dimensions and greater noise. For instance, applying MDL to healthcare data, social media datasets, or large-scale financial transactions can reveal the limitations and potential enhancements necessary to improve the algorithm's robustness. Future work could focus on adapting MDL to handle these new challenges, allowing for more accurate and efficient data compression and pattern recognition across diverse domains.

\subsubsection{Enhancing NLP for Broader Applicability}

Similarly, the use of NLP in this study is based on standard methods. However, there are significant opportunities for improving the performance of NLP by integrating domain-specific pre-trained models, such as those used in sentiment analysis or medical text mining. Additionally, hybrid approaches that combine NLP with deep learning techniques or ensemble methods could enhance the ability to process unstructured data, improving scalability and robustness. This would enable NLP to address more complex tasks, including those in areas with large, noisy, and unstructured datasets.

\subsection{Future Directions for Methodological Advancements}

Beyond domain-specific improvements, future work could involve the integration of MDL and NLP with other cutting-edge machine learning techniques. For example, combining MDL with reinforcement learning could help the algorithm adapt and evolve as new data becomes available, ensuring that it remains effective over time. Similarly, applying transfer learning techniques could allow both MDL and NLP to benefit from knowledge learned in one domain and apply it to others, increasing their efficiency and versatility.

The development of hybrid models that combine MDL, NLP, and emerging techniques such as graph-based methods or unsupervised learning could offer even greater flexibility and power, allowing for more comprehensive analysis of complex, real-world data.

\subsection{Conclusion}

This paper has demonstrated the applicability of MDL and NLP algorithms to a predefined dataset. However, there is considerable potential to extend these methods to new problem domains and datasets, and to innovate by combining them with other advanced techniques. These enhancements will provide a more robust, scalable, and adaptable approach to data analysis, opening new avenues for future research.

\section{Future Work}

There are several directions for advancing this research to improve both accuracy and computational efficiency in parsing date-time formats. One promising avenue is the development of a multi-format recognition mechanism that identifies cases where multiple date formats coexist within a single column. Such situations are common in datasets where multiple sources or entry methods are combined. Implementing format-switching predicates or conditional parsing rules would allow the system to dynamically apply different formats within the same data sequence, ultimately reducing error rates in these mixed-format datasets.

Another area of focus is extending the framework to handle numeric date representations, such as integer formats like \texttt{20241110} for \texttt{2024-11-10}. Numeric representations are often more computationally efficient to parse using arithmetic operations (e.g., modulo and division) rather than through locale-sensitive string parsing functions. Additionally, the limited number of possible numeric formats in date encoding could allow us to predefine a comprehensive set of parsing rules that cater specifically to this kind of data, significantly speeding up the parsing process in both real-time and batch-processing environments.

Addressing historical or locale-specific date formats could further broaden the applicability of this research. Many datasets spanning multiple decades or centuries contain date formats specific to a region or period, such as ordinal dates or Roman numerals in archaeological records. Developing lookup tables or probabilistic models to recognize and categorize these specific formats could improve both accuracy and utility for historical or regional datasets, where variations in date representation can significantly impact data interpretation.

Finally, there is potential to integrate these parsing techniques into database management systems (DBMS) and visualization platforms, providing a seamless experience for end-users who rely on clean, context-aware data for their analyses. Embedding these algorithms within visualization tools or relational database systems would enable users to interact with date-parsed data in real time, enhancing productivity without requiring users to leave the application or script custom parsing logic. Integrating MDL or NLP-based parsing directly into production tools would also allow the system to learn and adapt based on user-provided feedback or observed patterns, leading to more accurate parsing over time and potentially contributing to the automation of data preparation.
\subsection{Limitations and Future Improvements}

Despite the demonstrated efficacy of the MDL and NLP-based algorithms, certain limitations inherent to their design and implementation warrant further exploration. One notable limitation is the dependence of the MDL algorithm on the ICU date format, which restricts its adaptability to region-specific or historically uncommon date representations not explicitly covered by the ICU library. For example, unconventional date formats, such as those incorporating Roman numerals or ordinal indicators (e.g., \texttt{1st Feb 2020}), often require manual preprocessing or additional parsing logic. This constraint limits the method's utility in datasets derived from diverse global sources or archival records.

The NLP-based algorithm, while versatile in interpreting ambiguous patterns, is computationally intensive and susceptible to misinterpretation when faced with incomplete or noisy data. Its reliance on the Probabilistic Context-Free Grammar (PCFG) further adds complexity, as tuning the grammar to specific datasets can be challenging without substantial domain knowledge. Additionally, both algorithms struggle in environments where multiple date formats coexist within the same dataset column, leading to potential inaccuracies or elevated error rates.

These limitations can be addressed through several future improvements. Expanding the ICU library's date format coverage by incorporating user-defined patterns or community-driven updates could significantly enhance the MDL algorithm's adaptability. Similarly, optimizing the NLP algorithm through implementation in a high-performance language such as C++ or Rust, coupled with hardware acceleration techniques, may reduce computational overhead and enable its deployment in real-time systems.

Furthermore, integrating a multi-format recognition mechanism that leverages conditional parsing or ensemble methods could improve parsing accuracy in mixed-format scenarios. Employing machine learning models trained on diverse datasets to predict and adapt to new or uncommon formats represents another promising avenue for development. By addressing these challenges, future iterations of these algorithms can extend their applicability and robustness across broader use cases and data contexts.

\section{Conclusion}

This research has explored and evaluated two automated algorithms—Minimum Description Length (MDL) and Natural Language Processing (NLP)—for identifying and parsing diverse date-time formats in structured datasets. Our findings indicate that both algorithms successfully identify formats based on a limited number of samples, with MDL demonstrating superior speed and suitability for real-time applications, while NLP offers greater flexibility in interpreting more complex or ambiguous data patterns. By validating these algorithms on a large dataset, we have underscored the high degree of variability in date formats encountered in real-world contexts, highlighting the importance of robust and adaptable parsing solutions for data preparation.

The MDL-based algorithm, with its efficient runtime performance, is especially suitable for integration in interactive data visualization or cleaning applications, where minimal user intervention is desirable. Meanwhile, the NLP-based approach’s syntactic capabilities present an additional advantage for cases requiring interpretive flexibility, such as identifying formats with irregular or unconventional patterns. This distinction between the two approaches offers the potential for tailored applications depending on the computational and contextual requirements of specific use cases.

Ultimately, this research contributes to the broader field of data preparation by addressing the unique challenges of parsing inconsistent data formats. Our work not only demonstrates the feasibility of modular parsing techniques for diverse temporal data but also paves the way for more sophisticated data-wrangling tools that can adapt to user inputs, contextual cues, and patterns in the data. With these advancements, we hope to facilitate data preparation for researchers, analysts, and engineers, enabling them to leverage cleaner, more accurate data across various applications and industries.

\bibliographystyle{IEEEtran}
\bibliography{paper} 
\end{document}